\documentclass[runningheads]{llncs}
\usepackage{graphicx}
\usepackage{amssymb}
\usepackage{amsmath}

\begin{document}
\title{Vision Transformer Based Model for Describing a Set of Images as a Story}

\author{Zainy M. Malakan\inst{1,2}\orcidID{0000-0002-6980-0992} \and\\
Ghulam Mubashar Hassan\inst{1}\orcidID{0000-0002-6636-8807} \and\\
Ajmal Mian\inst{1}\orcidID{0000-0002-5206-3842}} 


\institute{The University of Western Australia, Perth WA 6009, Australia \\
\email{\{ghulam.hassan,ajmal.mian\}@uwa.edu.au}
\and
Umm Al-Qura University, Makkah 24382, Saudi Arabia\\
\email{\{zmmalakan\}@uqu.edu.sa}
}
\maketitle              
\begin{abstract}
Visual Story-Telling is the process of forming a multi sentence story from a set of images. Appropriately including visual variation and contextual information captured inside the input images is one of the most challenging aspects of visual storytelling. Consequently, stories developed from a set of images often lack cohesiveness, relevance, and semantic relationship. In this paper, we propose a novel Vision Transformer Based Model for describing a set of images as a story. The proposed method extracts the distinct features of the input images using a Vision Transformer (ViT). Firstly, input images are divided into 16X16 patches and bundled into a linear projection of flattened patches. The transformation from a single image to multiple image patches captures the visual variety of the input visual patterns. These features are used as input to a {\em Bidirectional-LSTM} which is part of the sequence encoder. This captures the past and future image context of all image patches. Then, an attention mechanism is implemented and used to increase the discriminatory capacity of the data fed into the language model, i.e. a {\em Mogrifier-LSTM}. The performance of our proposed model is evaluated using the Visual Story-Telling dataset (VIST), and the results show that our model outperforms the current state of the art models.

\end{abstract}
\section{Introduction}
Visual description or storytelling (VST) seeks to create a sequence of meaningful sentences to narrate a set of images. It has attracted significant interest from the vision to language field. However, compared to image \cite{li2022comprehending,fei2022deecap} and video \cite{pan2020spatio,seo2022end,lin2022swinbert} captioning, narrative storytelling \cite{9647213,kim2018glac,Liu_Fu_Mei_Chen_2017} has more complex structures and incorporates themes that do not appear explicitly in the given set of images. Moreover, describing a set of images is challenging because it demands algorithms to not only comprehend the semantic information, such as activities and objects in each of the five images along with their relationships, but also demands fluency in the phrases as well as the visually unrepresented notions.

\begin{figure}
\includegraphics[width=\textwidth]{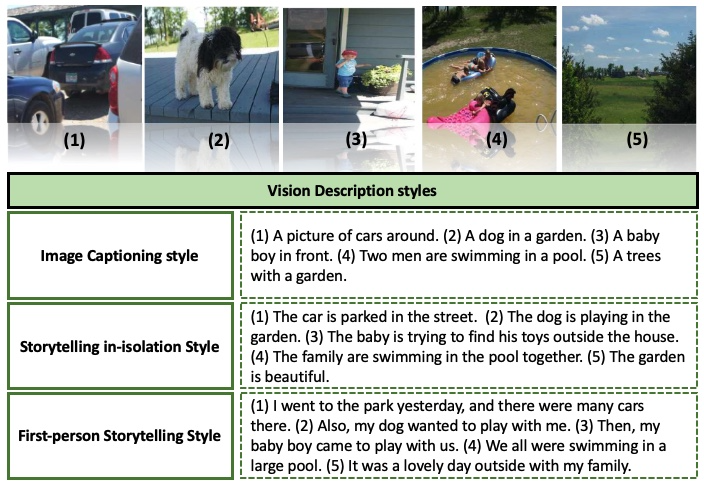}
\caption{An example of three vision description techniques includes a single picture caption, story-like caption, and narrative storytelling, which is our aim.} \label{fig1}
\end{figure}

Recent storytelling techniques utilise sequence-to-sequence (seq2seq) models \cite{visapp21,kim2018glac} to produce narratives based story on a set of images. The key idea behind these approaches is to implement a convolutional neural network (CNN), (\emph{i.e}, {\em a sequence encoder}), to extract the visual features of the set of images. Then, combining these visual features, a complete set of image representations is obtained. The next step is to input this representational vector into a hierarchical long-short-term memory (LSTM) model to form a sequence of sentences as a story. This approach has 
dominated this area of research owing to its capacity to generate high-quality and adaptable narratives.

Figure \ref{fig1} illustrates the technical challenges between single image captioning style, isolation style, and storytelling style for a set of five images. For example, the first sentence of all the three blocks in Figure \ref{fig1} annotations show the following: ``A picture of cars around.", ``The car is parked in the street.", and ``I went to the park yesterday, and there were many cars there.". The first description is known as image captioning style which conveys the actual and physical picture information. The second description is known as storytelling in-isolation style which catches the image content as well, but it is not linked to the following sentence. The final description is known as first-person storytelling style which explains more inferences about the image as a story-based sentence and also links to the subsequent sentence.

In order to solve the above challenges and difficulties, we propose a novel methodology that explores the significance of spatial dimension conversion and its efficacy on Vision Transformer (ViT) \cite{dosovitskiy2020image} based model. Our method proceeds by extracting the feature vectors from the given images by dividing them into 16X16 patches and feeding them into a {\em Bidirectional-LSTM} ({\em Bi-LSTM}). This models the visual patches as a temporal link among the set of images. By using the {\em Bidirectional-LSTM}, we represent the temporal link between patches in both forward and backward directions. To preserve the visual-specific context and relevance, we convert the visual features and contextual vectors from {\em Bi-LSTM} into a shared latent space using a {\em Mogrifier-LSTM} architecture \cite{melis2019mogrifier}. During the first layer's gated modulation, the initial gating step scales the input embedding based on the ground truth context, producing a contextualized representation of the input. This combination of multi-view feature extraction and highly context-dependent input information allows the language model to provide more meaningful and contextual descriptions of the input set of images.

\noindent The following is a summary of the contributions presented in this paper:
\begin{itemize}
\item We propose a novel ViT sequence encoder framework, that utilises multi-view visual information extraction for appropriate narrative based story on the given set of images as input.

\item We take into account the context of the past as well as the future and employ an attention mechanism over the contextualized characteristics that have been obtained from Vision Transformer (ViT) to construct semantically rich narratives from a language model.

\item We propose to combine {\em Mogrifier-LSTM} with enriched visual characteristics (patches) and semantic inputs to generate data-driven narratives that are coherent and relevant.

\item We demonstrate the utility of our proposed method through multiple evaluation metrics on the largest known Visual Story-Telling dataset (VIST) \cite{huang2016visual}\footnote{https://visionandlanguage.net/VIST/}. In addition, we compare the performance of our technique with existing state of the art techniques and show that it outperforms them on various evaluation metrics.

\end{itemize}

\section{Related Works}
This section presents a review of literature on different visual captions that directly relate to narrative storytelling techniques, followed by the literature on visual storytelling methods.

\subsection{Visual Understanding}
Visual understanding algorithms, which include image and video captioning, are the most significant sort of networks utilized to tackle the problem of narrative storytelling. Since it is most relevant to our study, we briefly discuss the recent literature on neural network-based image and video captioning. Typically, these models extract a vector of visual features using a CNN and then transmit this vector to a language model for caption synthesis. 
\subsubsection{Image Captioning (IC)}consists of a single frame (\emph{i.e}, an image) defined by a single phrase. Approaches may be further classified as rule-based methods \cite{do2020reference,mogadala2020integrating} and deep learning-based methods \cite{phukan2020efficient,huang2019attention}. The rule-based approaches apply the traditional methodology of recognizing a restricted number of pre-defined objects, activities and locations in the image, and describing them in natural language using template-based techniques. On the other hand, due to recent advances in deep learning, the vast majority of current methods are dependent on deep learning as well as scientifically advanced techniques such as attention \cite{wang2020learning}, reinforcement learning \cite{shen2020remote}, semantic attributes integration \cite{li2019entangled}, and modeling of subjects and objects \cite{ding2019image}. However, none of these algorithms are designed to produce a narrative-based description of a set or collection of images.

\subsubsection{Video Captioning (VC)}defined as multi-frame description that can explain many frames (\emph{i.e}, a video) in a single statement. VC and storytelling techniques are quite similar as they both utilize an encoder-decoder framework. The encoder is composed of a 2D/3D CNN that extracts visual information from a set of input frames. This information is subsequently converted into normal language phrases using a decoder or a language model based on either a recurrent neural network \cite{cho2014properties,pang2022video} or a transformer network \cite{zhou2018end,li2022long,jiang2022double}. Although VC methods can describe multi-frames in a single caption efficiently, it does not generate a story or multi-sentence descriptions for a given set of images.

\subsection{Storytelling Methods}
Telling a story based on a set of images is an easy task for humans, but an extremely difficult task for machines. Coherent, relevant, and grammatically correct sentences must be generated for a story-based description. For example, Park et al. (2015) \cite{park2015expressing} illustrated that using bidirectional recurrent neural network (BRNN) is more efficient than a usual recurrent neural network (RNN) because BRNN captures forward and backward image features, which enables the model to interact with the whole story's sentences. Similarly, Sequence-to-Sequence (Seq2Seq) techniques, which utilize CNN+Bi-LSTM \cite{kim2018glac}\footnote{https://github.com/tkim-snu/GLACNet} or CNN+GRU \cite{wang2018show} as an encoder and RNN as a decoder enhanced storytelling prediction from a set of images.

In addition, the concept of designing composite rewards as a strategy for storytelling problems was introduced \cite{hu2020makes}\footnote{https://github.com/JunjieHu/ReCo-RL}, which improved the natural flow of the generated story. A novel decoder-encoder framework using {\em Mogrifier-LSTM} \cite{visapp21} was also proposed to improve the coherence, relevance, and information of the generated story. Recently, the object detection technique using {\em YOLO}v5 \cite{9647213} is embedded with the encoder to improve the relevance of the story sentences.

Different from previous works, our proposed method derives characteristics from the multiple visual features (i.e., patch features) based on the human-like approach to generate stories. This helps to propose an approach that is both computationally efficient and capable of producing coherent, relevant, and informative stories.

\begin{figure}
\includegraphics[width=\textwidth]{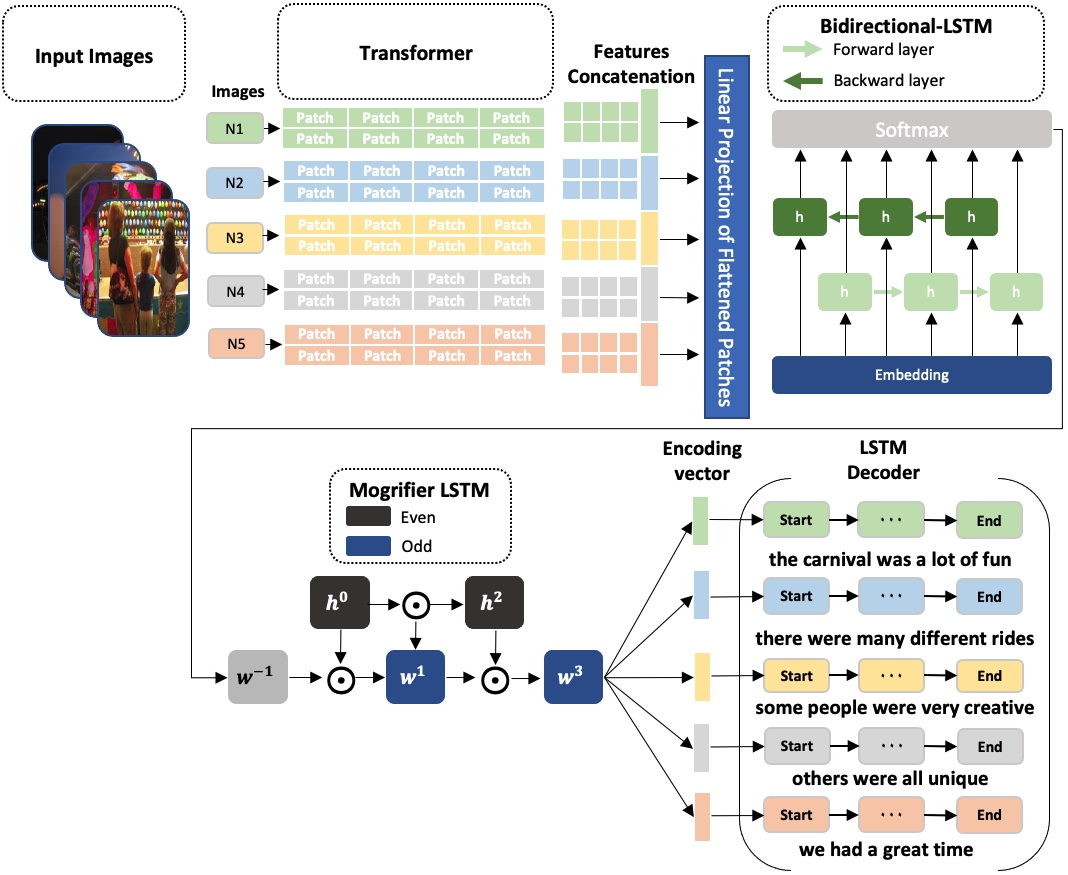}
\caption{An overview of our proposed model which consists of a sequence encoder and decoder. The sequence encoder process is implemented by both the Vision Transformer (ViT) and the {\em Bidirectional-LSTM}. The decoder process is performed by the {\em Mogrifier-LSTM} as well as the standard LSTM.} \label{method}
\end{figure}

\section{Proposed Method}
Figure \ref{method} presents the overall architecture of our proposed model which comprises of Vision Transformer (ViT), sequence encoder and decoder modules. In the first step, the image features are extracted using ViT, which divides each image into {\em16X16} patches and encodes them. Then all extracted image patch features are further encoded by {\em Bidirectional-LSTM} module which extracts the temporal context of the images. The connection between the sequence encoding and image features is captured by the Attention module on two levels: the patch level and the image-set patch level. Finally, the decoder module is responsible for the generation of a sequence of sentences as human story-like by making use of the {\em Mogrifier-LSTM} architecture. The following discussion delves into the specifics of the aforementioned three modules.

\subsection{Vision Transformer (ViT)}
A set of {\em I} images are fed by the data-loader as $I_s = (I_1, I_2, ..., I_s)$, where
\begin{equation}
 I = [ I_1, I_2, ..., I_N]~~s.t.~~ I_N \in
  \mathbb{R}^{H\times W \times C},
\end{equation}
$s \in \{1, 2, 3, 4, 5\}$ which is a set of five images with HxWxC (Height x Width x Channels) shape that presents a unique representation of storytelling from the dataset. To extract image features, we utilized Vision Transformer (ViT) \cite{dosovitskiy2020image} which breaks the given {\em I} image into {\em N} equal-sized, non-overlapping patches of shape (P, P, C) and linearly maps each patch to a visual representation. We define the extracted features as the combination of patches from the ViT model as follows:
\begin{equation}
 I_0 = [ I_p^1E; I_p^2E; ...; I_p^NE]~~s.t.~~ E \in
  \mathbb{R}^{(P^2.C) \times D}
\end{equation}
where {\em P} is the defined parameter as in grid order (left to right, up to down) while C represents the total number of channels. Then we flatten all patches which produces {\em n} line feature vectors of shape $(1, P^{2\star} C)$.
The patches that have been flattened are multiplied by a trainable embedding tensor of shape $(P^{2\star} C, D)$, which gains the ability to linearly project each flat patch to dimension D. As a result, we produce rich embedded patches of shape {\em n} = $ (1, D)\in \mathbb{R}^{(1, D)}$.

\subsection{Features Encoding}
The purpose of visual storytelling is first to comprehend the flow of events occurring in each image and then to produce a consistent narrative similar to how humans narrate a story. As a set of $P = P_1, P_2, ..., P_l$, where P represents the total number of image patches included in I as well as the number of corresponding contexts in each story.
In order to represent these relationship features, we utilize a {\em Bidirectional-LSTM}, which compiles the sequential information of $P$ patches in both {\em forward} and {\em backward} direction. Our sequence encoder requires an input of image feature vector $\boldsymbol f_i$ at every time step `$t$' where $i \in \{1, 2,.., 5\}$. Eventually, the sequence encoder part of the model encodes the whole image set, comprising all the image patches and provides contextual information $\boldsymbol h_{se} = [\overrightarrow{\boldsymbol h_{se}}; \overleftarrow{\boldsymbol h_{se}}]$ through the final hidden-state at time step number $t=5$.

\subsection{Story Generation}
Since modelling sequential inputs must lead to generating coherent sentences, the solution to the challenge lies in how well the model learns the context. This is particularly problematic for issues that need high levels of coherence and relevance. To solve this, we utilize the standard LSTM \cite{hochreiter1997long}, which forms the current hidden state denoted by $\boldsymbol h^{\langle t \rangle}$, based on the previous hidden state, represented by $\boldsymbol h_{prev}$, and refreshes its memory state $\boldsymbol c^{\langle t \rangle}$. Further, standard LSTM utilizes input gates $\boldsymbol \Gamma_i$, forget gates $\boldsymbol \Gamma_f$, and output gates $\boldsymbol \Gamma_o$ which are determined as follows:
\begin{equation}
\boldsymbol \Gamma_f^{\langle t \rangle} = \sigma(\boldsymbol M_f[\boldsymbol h_{prev}, \boldsymbol w_{t}] + \boldsymbol B_f),
\end{equation}
\begin{equation}
\boldsymbol \Gamma_i^{\langle t \rangle} = \sigma(\boldsymbol M_i[\boldsymbol h_{prev}, \boldsymbol w_{t}] + \boldsymbol B_i),
\end{equation}
\begin{equation}
\boldsymbol {\tilde c}^{\langle t \rangle} = \tanh{(\boldsymbol M_c[\boldsymbol h_{prev},  \boldsymbol w_{t}] + \boldsymbol B_c)},
\end{equation}
\begin{equation}
\boldsymbol c^{\langle t \rangle} = \boldsymbol \Gamma_f^{\langle t \rangle} \odot \boldsymbol c^{\langle t-1 \rangle} + \boldsymbol \Gamma_i^{\langle t \rangle} \odot \boldsymbol {\tilde c}^{\langle t \rangle},
\end{equation}
\begin{equation}
\boldsymbol \Gamma_o^{\langle t \rangle} = \sigma(\boldsymbol M_o[\boldsymbol h_{prev}, \boldsymbol w_{t}] + \boldsymbol B_o),
\end{equation}
\begin{equation}
\boldsymbol h^{\langle t \rangle} = \boldsymbol \Gamma_o^{\langle t \rangle} \odot tanh(\boldsymbol c^{\langle t \rangle})
\end{equation}
where $\boldsymbol w$ is the word vector embedded in the input at time step `$t$' (for simplicity, we eliminate $t$), $\boldsymbol M_*$ represents the transformation matrix that is learned at each state, $\boldsymbol B_*$ are the biases, $\sigma$ shows the logistic sigmoid function, and $\odot$ is the product of the vectors' Hadamard transform. In our generation module, the attention vector $\boldsymbol \zeta_i$ from the sequence encoder output is used to set up the LSTM hidden state $\boldsymbol h$.

Furthermore, we boost the standard LSTM functionality to generate more cohesive and relevant story-like sentences by integrating a {\em Mogrifier-LSTM} \cite{melis2019mogrifier}. The two inputs,~$\boldsymbol w$ and $\boldsymbol h_{prev}$, modulate each other in an odd and even fashion before being sent into the standard LSTM. In order to accomplish this goal, the {\em Mogrifier-LSTM} instead scales the columns of each of its weight matrices throughout $\boldsymbol M_*$ via {\em Mogrifier-LSTM} gated modulation. In formal terms, ~$\boldsymbol w$ is gated based on the previous step $\boldsymbol h_{prev}$ as gated input. A similar approach of gating prior time step output is used with the previous gated input.

Following the completion of five rounds of mutual gating, as recommended by Malakan \textit{et al.} \cite{9647213}, the most highly indexed versions of $\boldsymbol w$ and $\boldsymbol h_{prev}$ are subsequently fed into  the standard LSTM in the order shown in Figure \ref{method}. Therefore, it may also be stated as: {\em mogrification} $(\boldsymbol w, \boldsymbol c_{prev}, \boldsymbol h_{prev}) = LSTM(\boldsymbol w^{\uparrow}, \boldsymbol c_{prev}, \boldsymbol h_{prev}^{\uparrow})$ where $\boldsymbol w^{\uparrow}$ and $\boldsymbol h_{prev}^{\uparrow}$ are the most significant possible indexed for the LSTM inputs $\boldsymbol w^i$ and $\boldsymbol h_{prev}^i$ respectively. Mathematically,
\begin{equation}
    \boldsymbol w^i = 2 \sigma (\boldsymbol M_{xh}^i \boldsymbol h_{prev}^{i-1}) \odot \boldsymbol w^{i-2}, \hspace{4mm} \text{\rm for~odd~} i \in [1,2,...,r],
\end{equation}
\vspace{-3mm}
\begin{equation}
    \boldsymbol h_{prev}^i = 2 \sigma (\boldsymbol M_{hx}^i \boldsymbol w^{i-1}) \odot \boldsymbol h_{prev}^{i-2}, \text{\rm ~~for~even~} i \in [1,2,...,r],
\end{equation}
where Hadamard product is $\odot$, which $\boldsymbol w^{-1} = \boldsymbol w$, $\boldsymbol h^0_{prev} = \boldsymbol h_{prev} = \boldsymbol \zeta_i$ and $r$ represents the total number of mogrification rounds which is a mogrifier hyperparameter. In addition, the default standard LSTM configuration, with $r$ set to $0$, operates without gated mogrification at the input stage. The use of matrix multiplication with a constant of $2$ ensures that the resulting transformations of the matrices $\boldsymbol M_{xh}^i$ and $\boldsymbol M_{hx}^i$ are close to the identity matrix.

\subsection{Data Pre-processing And Model Training}
A vocab size of $6464$ was extracted from the Visual Story-Telling dataset (VIST) with a minimum word count threshold of 8. In addition, we used the size of $256$ as a dimension of word embedding vectors. Then, all VIST images are resized to $224$ $\times$ $224$ pixels from the original size and used as an input to the pre-trained Vision Transformer (ViT). For the training parameters, the Adam optimizer was used, and the learning rate was set at 0.001, while the weight decay was established at 1e-5. Also, a teacher-forcing strategy was utilized in our proposed model to help the model train faster. 

All of these settings were calibrated on our NVIDIA GPU, which has 12 GB of memory. For the maximum possible usage of available memory, the batch size was set to 8 during training. This ensured that we obtained the most out of the memory that was available to us. It's worth mentioning that greater GPU memory provides increased batch sizes, which assist the model to train faster. The model has been successfully trained for a total of 83 epochs utilizing about 390K steps. Finally, each epoch of our model was saved locally on our computer. Then, the optimum performance of the model was carefully chosen from epoch 59 since, after epoch 59, the model began to overfit the data and the loss error began to increase, resulting in decreased model accuracy.

\section{Experiments and Results}
First, we introduce the Visual Story-Telling dataset (VIST) used to evaluate our proposed model. Next, we discuss the results of our proposed model and compare them to other state of the art models. Finally, we give detailed analysis of a few cases in terms of the generated stories and the scores.

\subsection{Dataset} \label{datadet}
Visual Story-Telling dataset (VIST) \cite{huang2016visual}\footnote{https://visionandlanguage.net/VIST/} is the only publicly accessible dataset that we are aware for storytelling problems. It comprises 210,819 distinct images that can be found in 10,117 different albums on Flickr and is arranged in sets of five different images. Two types of stories accompany each set of images. One is called Description In Isolation (DII) and includes individual image descriptions that can be useful for research in image captioning. The second one is called Story In Sequence (SIS) which is more relevant to storytelling problems and comprises a whole paragraph in precisely five sentences representing a story. In all dataset statements, it is essential to note that the names of the individuals are adjusted by ``[male and female]", places by ``[location]", and organizations by ``[organization]".

\subsection{Performance Comparison}
Automatic evaluation metrics are the most common technique for estimating the effectiveness of the automatically generated story. Therefore, we validate our proposed model using automatic evaluation metrics, which also allows us to compare it to the current state-of-the-art methods. Table \ref{tableStateOfArt} displays the most recent frameworks used in storytelling challenges. These frameworks were published since 2018 and obtained promising results on the VIST dataset. We compare our proposed model using multiple evaluation metrics, which are: BLEU-1, BLEU-2, BLEU-3, BLEU-4, CIDEr, METEOR, and ROUGE-L. The script for computing the evaluation measures was released by \cite{hu2020makes}\footnote{https://github.com/JunjieHu/ReCo-RL}.

From the experiments, we observe that our model performs better than the state-of-the-art models on all of the given evaluation measures, except for the BLEU-1, BLEU-4, and CIDEr. Table \ref{tableStateOfArt} presents the results for all of the mentioned models sorted by the year in which they were released. Overall, our proposed model outperforms the compared models by 0.3 points in BLUE-2, 1.1 points in BLUE-3, 0.7 points in ROUGE-L and 0.2 points in METEOR.

\begin{table}[t]

\caption{A comparison of our proposed model with the recently published methods on the Visual Story-Telling dataset (VIST). Quantitative results were obtained using seven different automated measures of evaluation. ``-" indicates that the authors of the corresponding study did not publish the results. The higher scores represent higher accuracy and the results in bold represent the best scores.}
\label{tableStateOfArt}
\resizebox{\textwidth}{!}{\begin{tabular}{|l|l|l|l|l|l|l|l|}

\hline
\textbf{Model}  & \textbf{B-1} & \textbf{B-2} & \textbf{B-3} & \textbf{B-4} & \textbf{CIDEr} & \textbf{ROUGE-L} & \textbf{METEOR}  \\
\hline \hline
AREL 2018 \cite{wang2018no}& 0.536 & 0.315 & 0.173 & 0.099  & 0.038  & 0.286 & 0.352  \\
\hline
GLACNet 2018 \cite{kim2018glac}& 0.56 & 0.321 & 0.171 & 0.091 & 0.041  & 0.264 & 0.306  \\
\hline
HCBNet 2019 \cite{al2019hierarchical} & 0.59 & 0.348 & 0.191 & 0.105 & 0.051  & 0.274 & 0.34  \\
\hline
HCBNet(w/o prev. sent. attention) \cite{al2019hierarchical}& 0.59 & 0.338 & 0.180 & 0.097 & 0.057  & 0.271 & 0.332   \\
\hline
HCBNet(w/o description attention) \cite{al2019hierarchical} & 0.58 & 0.345 & 0.194 & 0.108 & 0.043  & 0.271 & 0.337  \\
\hline
HCBNet(VGG) 2019 \cite{al2019hierarchical} & 0.59 & 0.34  & 0.186 & 0.104 & 0.051 & 0.269 & 0.334   \\
\hline
ReCo-RL 2020 \cite{hu2020makes} & -& - & - & 0.124 & 0.086 & 0.299 & 0.339   \\
\hline
BLEU-RL 2020 \cite{hu2020makes} & -& - & - & 0.144 & 0.067 & 0.301 & 0.352  \\
\hline
VS with MPJA 2021 \cite{guo2021steganographic} & 0.601 & 0.325 & 0.133& 0.082 & 0.042  & 0.303  & 0.344 \\
\hline
CAMT 2021 \cite{visapp21} & 0.64 & 0.361 & 0.201  & \textbf{0.184} & 0.042  & 0.303  & 0.335 \\
\hline
Rand+RNN 2021 \cite{chen2021commonsense} & - & - & 0.133  & 0.061 & 0.022  & 0.272  & 0.311 \\
\hline
SAES Encoder-Decoder OD 2021 \cite{9647213}  & 0.64 & 0.363 & 0.196  & 0.106 & 0.051  & 0.294 & 0.330  \\
\hline
SAES Encoder-Decoder OD \& Noun 2021 \cite{9647213}   & 0.63 & 0.357 & 0.195 & 0.109 & 0.048 & 0.299 & 0.331  \\
\hline
SAES Encoder OD 2021 \cite{9647213}  & \textbf{0.65} & 0.372 & 0.204  & 0.12 & \textbf{0.054}  & 0.303 & 0.335  \\
\hline \hline
Our Proposed Model  & 0.63 & \textbf{0.375} & \textbf{0.215}  & 0.123 & 0.044  & \textbf{0.310} & \textbf{0.354}  \\
\hline
\end{tabular}}

\end{table}

\begin{figure}
\includegraphics[width=\textwidth]{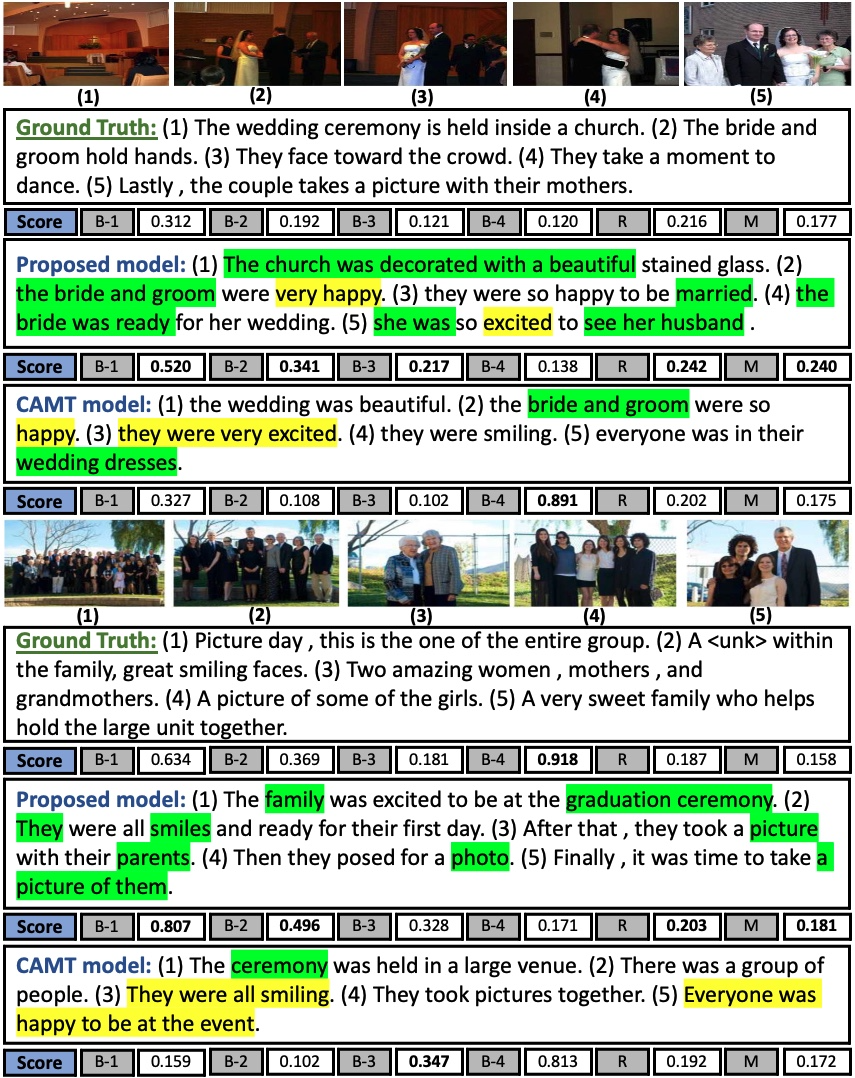}
\caption{Examples of our generated stories in comparison to the CAMT model \cite{visapp21} and ground truth. Text highlighted in green indicates high relevance to the image/story, while text highlighted in yellow means that it is not highly relevant but instead contains general information. BLEU-1 (B-1), BLEU-2 (B-2), BLEU-3 (B-3), BLEU-4 (B-4), ROUGE (R), and METEOR (M) scores are shown below each story, with bold scores indicating the highest value.} \label{exp2}
\end{figure}

\subsection{Storytelling Example Analysis}
Automatic metrics are not a perfect reflection of the accuracy of the stories. Therefore, we conducted an in-depth analysis of the stories produced by our proposed model and the ground truth. In addition, we compared our stories with stories produced by the recently proposed CAMT 2021 model \cite{visapp21}. 

Figure \ref{exp2} illustrates two different stories from a set of five images from our proposed model, followed by the stories that were Generated using CAMT. The highlighted text in green shows parts that are highly relevant to the story, while the highlighted text in yellow indicates less relevant or general information that is not obvious from the images. We also report the automatic evaluation metrics below each story in Fig. \ref{exp2}.

\subsubsection{Text Generation Analysis:}Both selected models have shown a persuasive example of a narrative that is representative of how humans write a story, and both of these examples are captivating. In contrast to CAMT model, our proposed model is able to extract more useful information from each input image. For instance, the 3rd sentence in the first scenario shows more relevance to the story, i.e. ``they were so happy to be married," as compared to CAMT model which predicted a less relevant sentence ``they were very excited." In addition, we noticed that the third and the last sentences in the second examples, which are ``They were all smiling" and ``Everyone was happy to be at the event."; the two sentences do not relate to the image itself in any manner, and the information they provide seems to be generic and applicable to many images. On the other hand, our proposed model generated a story that was more logically consistent with the story and relevant to the images.

\subsubsection{Generated Story Scores:}It is essential to demonstrate the model's performance in contrast with traditional automated evaluation metrics. Each set of images comes with a total of five different stories that were written by real people (\emph{i.e.}, ground truth), as mentioned in section \ref{datadet}. One of these stories was extracted randomly and removed from the collection. Next, we compared the story generated by our proposed model, CAMT model and the removed ground truth story (we name this Human Generated Story or HGS) with the rest of the four stories in the collection of VIST dataset. In BLEU-1, we found that our proposed model obtains the highest score on the second example, with almost 0.17 points more than HGS and 0.65 points more than the CAMT model; in BLEU-2, our generated story obtains over 0.14 points more than HGS and CAMT model; in BLEU-3, our model obtains 0.9 points more than the HGS and CAMT model in the first example, but the CAMT model receives 0.2 points more than ours and 0.16 points over the HGS in the second example; in BLEU-4, CAMT obtains almost 0.77 points more than our model and the HGS in the first example, while the HGS receives 0.74 points more than our model and 0.10 points over the CAMT model in the second example; in ROUGE-L, we reported that our model obtains almost 0.2 points more than both the HGS and the CAMT model in both examples; and in METEOR, we find that our model obtains almost 0.7 points more than both the HGS and the CAMT model in both examples. On the other hand, the performance of our proposed model is sufficiently high across practically all of the automated evaluation metrics, with the exception of BLEU-4, as is shown through Figure \ref{exp2}.

\section{Conclusion}
This article presented a novel storytelling approach for describing a set of images in a coherent manner. Our proposed framework is robust, which consists of a sequence encoder that receives multi-view image patches from Vision Transformer (ViT) as an input to a {\em Bidirectional-LSTM}, and a decoder with a standard LSTM enhanced by {\em Mogrifier-LSTM} that has five rounds of mogrifircation. Furthermore, we utilize an attention mechanism that enables our model to capture a specific significant context in response to a particular visual area while still keeping the more significant story context in mind. We found that our proposed model performs better on most of the automatic evaluation metrics than current state-of-the-art approaches except for BLEU-1, BLEU-4 and CIDEr scores. Additionally, we presented a comprehensive analysis of multiple examples, which indicated that our generated stories are more relevant and coherent.

\subsubsection*{Acknowledgments:}
This research received complete funding from the Australian Government, which was sponsored through the Australian Research Council (DP190102443).

\bibliographystyle{splncs04}
\bibliography{egbib}

\end{document}